%% file: formatting-instructions-latex.tex
\documentclass[letterpaper]{article}
\usepackage{aaai}
\usepackage{times}
\usepackage{helvet}
\usepackage{courier}
\usepackage{times}
\usepackage{latexsym}

\usepackage[T1]{fontenc}

\usepackage[utf8]{inputenc}

\usepackage{microtype}
\usepackage{graphicx}
\usepackage{bbm}
\usepackage{amssymb}
\usepackage{amstext}
\usepackage{booktabs}
\usepackage{multirow}
\usepackage{tabularx}
\usepackage{float}
\usepackage{caption}
\usepackage{amsmath}
\usepackage{url}
\usepackage{marvosym}

\begin{document}
%
\title{Exploring LLM-based Data Annotation Strategies for Medical Dialogue Preference Alignment}
\author{
Chengfeng Dou\textsuperscript{1,2}, 
Ying Zhang\textsuperscript{3},
Zhi Jin\textsuperscript{1,2}\Letter, 
Wenpin Jiao\textsuperscript{1,2}\Letter, 
Haiyan Zhao\textsuperscript{1,2}, \\
\textbf{Yongqiang Zhao\textsuperscript{1,2}}, 
\textbf{Zhengwei Tao\textsuperscript{1,2}}\\
\textsuperscript{1} School of Computer Science, Peking University;\\ 
\textsuperscript{2} Key Laboratory of High Confidence Software Technologies(PKU), MOE, China \\
\textsuperscript{3} Beijing Key Lab of Traffic Data Analysis and Mining, Beijing Jiaotong University, Beijing, China\\
\texttt{\{chengfengdou,zhijin,jwp,zhhy.sei\}@pku.edu.cn}\\
~\texttt{\{tttzw,yongqiangzhao\}@stu.pku.edu.cn} \texttt{\{19112043\}@bjtu.edu.cn}
}

\maketitle
\input{chapter/00_abs}
\input{chapter/01_intro}
\input{chapter/02_related_work}
\input{chapter/03_exp_setup}

\input{chapter/04_methodology}

\input{chapter/05_exp_result}
\input{chapter/06_discussion}

\input{chapter/07_conclusion}

\bibliography{myfile}
\bibliographystyle{aaai}


\end{document}

%% file: chapter/00_abs.tex
\begin{abstract}
This research examines the use of Reinforcement Learning from AI Feedback (RLAIF) techniques to improve healthcare dialogue models, with the aim of tackling the challenges of preference-aligned data annotation while reducing the reliance on medical experts. We argue that the primary challenges in current RLAIF research for healthcare are the limitations of automated evaluation methods and the difficulties in accurately representing physician preferences. To address these challenges, we present a new evaluation framework based on standardized patient examinations. This framework is designed to objectively assess the effectiveness of large language models (LLMs) in guiding users and following instructions, enabling a comprehensive comparison across different models. Furthermore, our investigation of effective ways to express physician preferences using Constitutional AI algorithms highlighted the particular effectiveness of flowcharts. Utilizing this finding, we introduce an innovative agent-based approach for annotating preference data. This approach autonomously creates medical dialogue flows tailored to the patient's condition, demonstrates strong generalization abilities, and reduces the need for expert involvement. Our results show that the agent-based approach outperforms existing RLAIF annotation methods in standardized patient examinations and surpasses current open source medical dialogue LLMs in various test scenarios.
\end{abstract}

%% file: chapter/01_intro.tex
\section{Introduction}
Sure, here's a paraphrased version of the given text:
In recent years, the use of large language models (LLMs) for generating medical dialogues has grown substantially~\cite{huatuogpt-2023,bao2023disc,zhu2023ChatMed,ullah2024challenges}. The performance of these models is heavily dependent on the availability of high-quality training data, steering current research towards creating comprehensive training datasets to improve model accuracy. Traditional studies typically gather data from online consultations or publicly accessible databases~\cite{yang2023zhongjingenhancingchinesemedical,han2023medalpacaopensourcecollection,yuan2024continued}. Recently, to tackle the issue of limited data for instruction tuning in the medical sector, various studies~\cite{xiong2023doctorglmfinetuningchinesedoctor,huatuogpt-2023,long2024llmsdrivensyntheticdatageneration,tang2023doessyntheticdatageneration,wu2023medical} have promoted using advanced LLMs like ChatGPT to produce synthetic medical data. These initiatives have yielded significant results, suggesting that synthetic data have great potential to improve medical dialogue models.

\begin{figure}[htbp]
    \centering
    \includegraphics[width=0.48\textwidth]{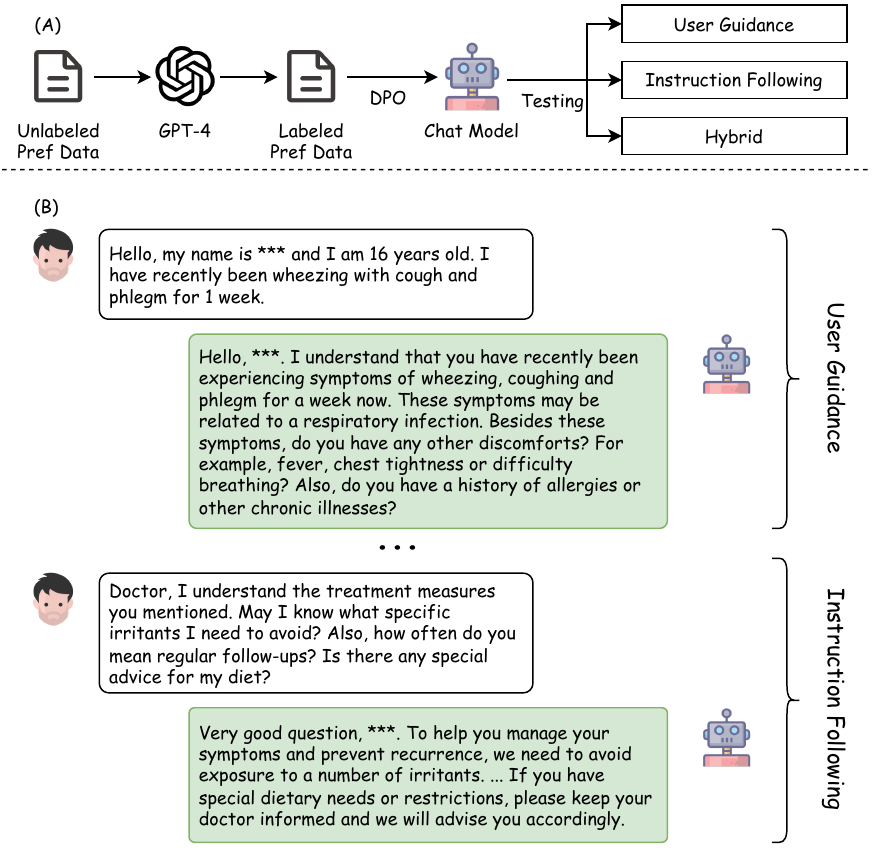}
    \caption{(A): The research method in this work.  (B): Two facets of dialogue competence evaluation. 'User Guidance' competence denotes the doctor's skill in prompting the patient to describe symptoms. Patients, often lacking medical knowledge, might fail to relay their information clearly and hence, need guidance from the doctor. 'Instruction Following' competence pertains to the doctor's skill in responding to the patient's questions accurately and amicably.}
    \label{fig:cover}
\end{figure}

Although many methods for data synthesis have been proposed for the instruction fine-tuning stage, generating preference alignment data at this stage still requires expert engagement~\cite{yang2023zhongjingenhancingchinesemedical,huatuogpt-2023} and incurs considerable costs. Reinforcement Learning with AI Feedback (RLAIF) has emerged as a promising alternative. This method suggests that Artificial Intelligence, particularly LLMs, can replace human annotators. Given the sophisticated capabilities and consistency of LLMs, RLAIF has been shown to be as effective as Reinforcement Learning with Human Feedback. This method has been used successfully to create concise summaries~\cite{lee2023rlaif} and reduce toxicity in conversations~\cite{bai2022constitutional}.
To our knowledge, the use of RLAIF for medical dialogue generation has not been extensively studied, highlighting the need to explore its feasibility for this specific application.

In our investigation of the application of RLAIF in medical dialogues, we identified two primary challenges. First, the reliability of current automated assessment systems is inadequate. Traditional reference-based methods struggle to accurately evaluate LLM outputs, as evidenced by previous research~\cite{dou2023plugmed,Ji2022,sas}. To mitigate this problem, recent studies~\cite{yang2023zhongjingenhancingchinesemedical,bao2023disc,huatuogpt-2023} often utilize GPT-4 to assess LLM responses. However, comparing different preference alignment approaches is rendered complex due to the marked deviation of GPT-4 preferences from those of human physicians and the significant variability in evaluation criteria across studies. Second, the effectiveness of RLAIF significantly depends on the prompt design, which needs to precisely capture doctors' preferences. This is challenging because doctors' strategies can vary greatly depending on the patient's condition. Conventional optimization objectives such as fluency and safety are too simplistic to address these differences.

To address the aforementioned challenges, we introduced a comprehensive solution. Initially, we developed an assessment framework, as illustrated in Figure~\ref{fig:cover}. Initially, we create an unlabeled preference dataset and prepare a dialogue model without  preference alignment. Subsequently, we label the preference dataset with various annotation strategies and use these labeled data to align the dialogue model's preferences employing the DPO algorithm. Lastly, we compare the benefits and drawbacks of different annotation strategies by evaluating the model across a range of downstream tasks focusing on user-guided, instruction-following, and combined capabilities. Mean while, using a standardized patient testing methodology, prevalent in the medical field, we established an objective evaluation system. This system assesses model proficiency in guiding users and following instructions, enabling broad comparability across models.

To enhance the optimization of RLAIF, we utilize a methodology termed Constitutional Artificial Intelligence (CAI)~\cite{bai2022constitutional}. This method allows for the creation of rules that represent doctors' preferences and aids us in exploring these preferences. Our research shows that doctors' preferences can be illustrated through flowcharts, which can then be converted into specific rules for annotating preference data. By combining CAI technology with flowchart-based representations of physician dialogue preferences, we have observed improvements in the model's diagnostic accuracy. 

To further minimize manual efforts, such as dialogue flow formulation, while improving the generalization of preference alignment, we propose a multi-agent approach for preference data annotation based on existing CAI findings. Unlike CAI, this method can autonomously create various medical dialogue flows based on the patient's condition, demonstrating better generalization with less expert input. Our experiments show that the agent-based approach significantly outperforms other RLAIF annotation methods in standardized patient testing and exceeds current open-source medical dialogue LLMs in multiple test sets.

In summary, our contributions are as follows.
\begin{itemize}

\item We have created a novel and thorough evaluation framework that precisely assesses the effectiveness of different preference alignment strategies in medical dialogue generation. This framework examines both the adherence to instructions and the efficiency in guiding users.

\item By formulating and combining multiple annotation guidelines, we have discovered that physicians' preferences can be depicted using flowcharts. Further, we propose a CAI-based approach to align physician preferences using flowcharts.

\item We present a multi-agent-based approach for annotating preference data, which demonstrates strong generalization abilities and greatly reduces the dependence on expert input for medical data annotation. Our experiments show that this agent-based method holds great potential for use in medical dialogue systems.

\end{itemize}

We recognize that an earlier version of this paper~\cite{dou2024integratingphysiciandiagnosticlogic} was presented in the ACL2024 Findings. The first publication concentrated exclusively on the CAI application, neglecting an in-depth exploration of RLAIF's role in medical dialogues. This manuscript expands the discussion of other RLAIF methods and improves the assessment of medical dialogue analysis.



%

%% file: chapter/02_related_work.tex
\section{Related Work}
\paragraph{Medical LLMs}
Significant advances have been made in medical dialogue models since the introduction of ChatGPT~\cite{chatgpt}. Recent research has primarily focused on creating large and high-quality instruction fine-tuning datasets for LLMs. Studies such as DoctorGLM~\cite{xiong2023doctorglm}, BenTsao~\cite{wang2023huatuo}, and ChatMed~\cite{zhu2023ChatMed} have proposed the utilization of powerful LLMs like ChatGPT to generate dialogue and question answering data at a low cost. However, ensuring the quality of data generated through this approach is challenging due to ChatGPT's tendency to generate incorrect information. To address this issue, Huatuo~\cite{huatuogpt-2023} suggested incorporating parts of real data into the generated data. Furthermore, to improve the readability of the real dialogue data, Huatuo refined it with ChatGPT, and this method has been widely adopted in subsequent research.
In addition to the dialogue data, several studies have aimed to generate various auxiliary task data. For example, DISC-MedLLM\cite{bao2023disc} and ClinicalGPT\cite{wang2023clinicalgpt} have integrated knowledge graph-related data into the training data to enhance the model's ability to answer common-sense questions. ClinicalGPT has also attempted to improve the diagnostic capacity of the model by including data from electronic medical records and medical examinations in training data. Although significant progress has been made in fine-tuning medical LLM instructions, there is still limited research on the preferred learning stage.

\paragraph{Preference Learning}
Preference alignment is a prominent focus in large model training research, as preferentially aligned models exhibit enhanced generalization ability in zero shot scenarios~\cite{kirk2023understanding}. 
Currently, the most renowned method for preference alignment is reinforcement learning from human feedback, which involves the utilization of four models for training. 
However, this approach has drawbacks, e.g., high engineering complexity and unstable training. 
In an effort to streamline the preference learning process, \cite{dpo} introduced a direct preference optimization algorithm that can bypass the need to train the reward model. 
Similarly, \cite{gulcehre2023reinforced} has proposed a self-reinforcement learning approach that uses the EM algorithm to eliminate the training of the critic model. 
Furthermore, RRHF\cite{yuan2023rrhf} suggests using learning rankings to replace reinforcement learning, thus strengthening the stability of learning. 
Furthermore, some initiatives, such as RLAIF\cite{lee2023rlaif}, aim to leverage AI to substitute manual preference data annotation, thus reducing annotation costs. 
Moreover, \cite{bai2022constitutional} proposes training constitutional evaluation models for self-reflection. 
\cite{sun2023salmon} has put forward the idea of using the Principle-following reward model as a replacement for the traditional reward model to achieve dynamic adaptation to human preferences.
Compared to these methods, our main contribution is to propose a preference learning approach for multiple rounds of dialogue.

%% file: chapter/03_exp_setup.tex
\section{Research Methodology}
In this section, we present the research framework for this study. We begin by outlining the task and then offer a detailed explanation of the data preparation, model setup, and evaluation techniques relevant to the specified task.

\subsection{Problem Definition}
We define an unlabeled preference dataset as a collection $\mathcal{D} = \langle H, C_1, F_1, C_2, F_2 \rangle$, where $H$ represents the context of the dialogue. Here, $C_1$ and $C_2$ are two potential responses to this context, while $F_1$ and $F_2$ signify the subsequent dialogue content that might follow if $C_1$ and $C_2$ are chosen, respectively. In contrast, labeled data are represented as $\mathcal{D}' = \langle H, A, R \rangle$, where $A$ and $R$ denote accepted and rejected responses, respectively. The objective of our study is to identify the most effective method, $f: \mathcal{D} \rightarrow \mathcal{D}'$, for providing feedback. This aims to train a model using $\mathcal{D}'$ to achieve optimal performance in the test set.

To investigate this issue, we prepare unlabeled data, initialize the model without preference alignment, and develop evaluation tasks. The preparation of the data and models is detailed in the following two subsections, and the evaluation methodology for the experiment is described in Section~\ref{sec:eval}.

\subsection{Data Preparation}
\label{sec:dp}
The primary objective of this study is to enhance the performance of the model in multiple conversations. This involves examining how current responses influence subsequent interactions, thereby enabling a comprehensive evaluation of the model's response quality. To achieve this, we construct a preference dataset for annotation that includes both potential candidate responses and projected future interactions, here termed future dialogue trajectories. We employ sampling and generation techniques to assemble this dataset.

\paragraph*{Sampling}
Data extraction from existing conversations offers a straightforward method to acquire potential responses. We utilize random sampling to select instances from the MedDialogue dataset \cite{zeng2020meddialog}. For each instance, one of the doctor's responses is randomly selected as a candidate response. The segment of the conversation preceding this response is defined as the dialogue history, and the segment following it is defined as the future dialogue trajectory.

\begin{figure}[th]
    \centering
    \includegraphics[width=0.48\textwidth]{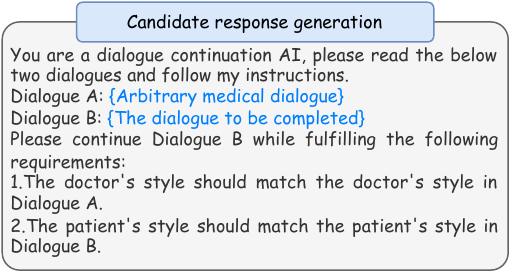}
    \caption{Prompt words used to generate candidate responses.}
    \label{fig:cg}
\end{figure}

\paragraph*{Generation}
Given that the sampling methodology is limited to producing a single candidate response for each dialogue history, we adopt a generative strategy to formulate alternative responses and potential future dialogue trajectories. Using ChatGPT according to the prompt described in Figure~\ref{fig:cg}, we execute this task. In this prompt template, by adjusting the dialogue example~(Dialogue A in the Figure~\ref{fig:cg}), we are able to influence the style of responses generated by ChatGPT, allowing the creation of various future potential scenarios.

Ultimately, we obtain 4,000 unlabeled preference data entries. For each candidate response, we preserve three rounds of future conversation traces for further analysis.

\subsection{Model Preparation}
\label{sec:backbone}
Considering the widespread adoption of Baichuan2-7B~\cite{yang2023baichuan} as the main model among open source Chinese LLMs in the healthcare domain, we select this model for our study to ensure a fair comparison. Furthermore, we incorporate the Qwen-7B~\cite{bai2023qwen} model in our control experiments to extend our analysis. For instruction fine-tuning purposes, we use the DISC-MedLLM~\cite{bao2023disc} dataset, which includes a broad spectrum of data. It contains a multi-round dialogue dataset MedDialog~\cite{zeng2020meddialog}, a single-round dialogue dataset cMedQA2~\cite{zhang2017chinese}, and the Knowledge Graph Q\&A dataset CMeKG~\cite{byambasuren2019preliminary}. To further enhance the medical diagnostic capabilities of our model, we integrate the MedMcQA~\cite{pal2022medmcqa} dataset, which comprises questions from the Medical Licensing Exam. Our upcoming experiments focus on models trained with the aforementioned datasets to investigate effective strategies for annotating preference data.

\subsection{Test Sets}
\label{sec:eval}
\begin{table}[ht]
\centering
\caption{Assessment tendencies of different test sets.}
\begin{tabular}{lcc}
\toprule
Dataset     & Testing Form      & Testing tendency \\ \midrule
CSPT & Simulation     & User Guidance    \\
WebMedQA & Single-Round        & Instruction Following \\
Meddg & Multi-Round  & Hybrid    \\
IMCS & Multi-Round    & Hybrid   \\ 
\bottomrule
\end{tabular}
\label{tab:datasets}
\end{table}

Currently, the healthcare dialogue sector is hindered by a lack of reliable assessment frameworks. Traditional evaluation metrics are insufficient, prompting substantial efforts to measure the effectiveness of dialogue models using predefined criteria such as fluency~\cite{yang2023zhongjingenhancingchinesemedical}, professionalism~\cite{bao2023disc}, and proactivity~\cite{wang2023cmb}. Nonetheless, the discrepancies in assessment standards used across different activities, coupled with the subjective aspect of developing these criteria, impede the impartial comparison of various labeling approaches.

To address the identified issues, this study suggests that the model be evaluated from the perspective of the physician's responsibility in patient communication. These responsibilities include diagnosing health conditions and responding to patient inquiries. For accurate diagnoses, physicians must obtain comprehensive information on symptoms from patients. Similarly, effective responses to inquiries require understanding the patient's intent and providing friendly and knowledgeable feedback. Drawing inspiration from these responsibilities, this research aims to assess the model's proficiency in ``user guidance'' and ``instruction following''. Given the frequent alternation between these tasks in real-world interactions, evaluating the model's ``hybrid'' capability is also important.

We utilize four different datasets in our benchmark: CSPT\footnote{\url{https://github.com/Chengfeng-Dou/SpTesting}}, WebMedQA~\cite{he2019applying}, Meddg~\cite{liu2022meddg}, and IMCS~\cite{imcs}, with their testing propensities presented in Table~\ref{tab:datasets}. The following sections demonstrate in detail the use of these test sets.

\begin{table}[ht]
\caption{
Statistics for the CSPT dataset. Patient Information provides details about the patient, while QA pairs consist of doctor-patient interactions in dialogue scripts, which aid in creating simulated patients. Major Symptoms, Major Medical Test, and Diseases are leveraged to assess the model's dialogue capabilities.
}
\label{tab:sp_testing}
\centering
\begin{tabular}{@{}lll@{}}
\toprule
Statistic & Item & Value \\ \midrule
\multirow{2}{*}{Count} & Department Num & 5 \\
 & Case Num & 72 \\ 
 \midrule
Avg Length & Patient Info & 493.3 \\ \midrule
\multirow{4}{*}{Avg Num} & QA pairs & 37.3 \\
 & Major Symptoms & 7.3 \\
 & Major Medical Test & 2.8 \\
 & Diseases & 1.7 \\ \bottomrule
\end{tabular}
\end{table}

\subsubsection{User Guidance}
Standardized patient testing (SPT) is frequently used by medical schools to evaluate the user-guidance abilities of trainee doctors. Here, we introduce this test method into LLM evaluation. During SPT, a volunteer plays the role of a standardized patient, trained to precisely replicate the symptoms and responses of a real patient. Throughout the test, the standardized patient does not provide information voluntarily; instead, the doctor must skillfully guide the patient to verbally express his symptoms to ensure an accurate diagnosis. The doctor's proficiency is ultimately evaluated based on whether essential symptom collection rates are achieved, necessary medical tests are recommended, and an accurate diagnosis is made.

The emergence of LLM has facilitated the implementation of computer simulations for standardized patients. Multiple prior studies~\cite{huatuogpt-2023,wei-etal-2018-task} have aimed to assess model performance using comparable methods. However, these studies often provide limited information on symptoms for the modeled patients, which typically impair the model's capability to faithfully simulate the patient's role. In the medical field, the conventional approach to addressing this is to have standardized patients learn a comprehensive dialogue script crafted by an expert, which authentically portrays the patient's responses to various questions. During an examination, the patient can respond to the doctor's inquiries according to the script, thereby ensuring the quality of the responses.

\begin{figure}
    \centering
    \includegraphics[width=0.48\textwidth]{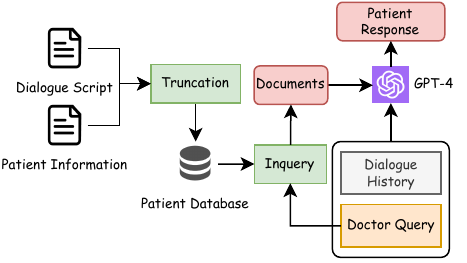}
    \caption{Structure of the Patient Simulator. The dialogue content and patient details are initially divided into smaller segments, which are then stored in individual databases for each patient. The text-embedding-ada-002 model from OpenAI is used to encode these segments, producing vectors for similarity retrieval. During operation, the simulator retrieves the top four most relevant segments from the database based on the doctor's inquiries. Subsequently, GPT-4 utilizes the prompt depicted in Figure~\ref{fig:ps_prompt} to formulate the patient's response by merging the dialogue history with the retrieved information.}
    \label{fig:patient_simulator}
\end{figure}

\begin{figure}
    \centering
    \includegraphics[width=0.48\textwidth]{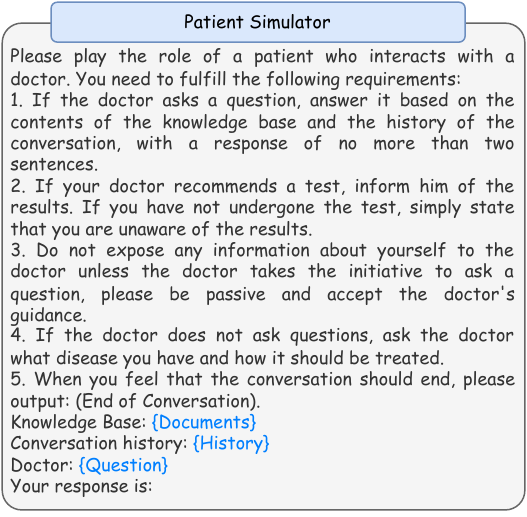}
    \caption{Prompt words for patient simulator, information drawn from the patient database is located at {Documents}, with the dialogue history and the doctor's inquiries placed at {History} and {Question}, respectively. {Question} serves to retrieve patient information. Notably, as {Question} includes sufficient context to ensure accurate retrieval, the last two rounds of dialogue are used as {Question}.}
    \label{fig:ps_prompt}
\end{figure}

Drawing inspiration from this, we develop the Chinese Standardized Patient Test (CSPT) dataset. To simulate patient scenarios, we use a retrieval-enhanced generation method that incorporates patient descriptions and dialogue scripts into the database, as illustrated in Figure~\ref{fig:patient_simulator}. To prevent the test from going on endlessly, we set the upper limit of the number of simulated dialogue rounds to five. The objective of standardized patient testing is to assess the comprehensiveness of the information collected about critical symptoms, the thoroughness of the recommendations for physical examinations, and the precision of disease diagnosis. These three dimensions of evaluation metrics are classified as symptoms (Sym.), testing (Test), and diagnosis (Dis.). Hence, we offer a set of checklists for every case. To ensure the accuracy of the results, we manually score the models against the checklist at the end of the simulated dialogue. Statistics related to the standardized patient test are presented in Table~\ref{tab:sp_testing}.

\subsubsection{Instruction Following}
Since the SPT is unable to directly evaluate the model's performance when dealing with patient inquiries, it is essential to include conventional dialogue evaluation techniques in the evaluation framework. For this purpose, we use WebMedQA~\cite{he2019applying}. This dataset consists of single round Q\&A interactions, where the patient's information is assumed to be comprehensive, and the doctor is expected to respond solely based on the information provided. Hence, this dataset is highly appropriate for assessing the model's instruction-following capability.

However, recent research~\cite{dou2023plugmed} has shown that traditional evaluation metrics such as BLEU and Rouge are not reliable to assess model performance. The main reason for this lack of reliability is that the responses generated by LLMs tend to be much longer than the reference answers. Current medical LLMs aim to provide comprehensive responses with more detailed information than reference responses, which causes their performance to be often undervalued when evaluated using metrics focusing on information precision.

In our study, we suggest evaluating LLMs with Rouge-L-Recall~(R@L), a metric that quantifies the recall rate. Given the limitation of the R@L metric in gauging semantic similarity, we introduce a new evaluation metric named GPT-Distance~(GPD). This metric checks the consistency of LLM output with reference answers. Specifically, we use GPT-4 to determine whether a prediction implies the reference answer, categorizing it as unimplied, partially implied, or fully implied. The prompt used is shown in Figure~\ref{fig:gpt_eval_prompt}. After obtaining all predictive classifications, the GPD is calculated with the formula $(2 \times |\text{Not}| + |\text{Partially}|) / |\text{ALL}|$. Here, $|\text{Not}|$ and $|\text{Partially}|$ denote the counts of samples categorized as ``Not'' and ``Partially'', respectively. $|\text{ALL}|$ signifies the total number of test samples.

\subsubsection{Hybrid}
We utilize two multi-round dialogue datasets, Meddg~\cite{liu2022meddg} and IMCS~\cite{imcs}, to assess the ``hybrid'' capability. These datasets, derived from actual doctor-patient interactions, allow for a thorough evaluation of the model's proficiency in both guiding the user and adhering to instructions. For evaluation, we continue to employ the metrics used in WebMedQA~\cite{he2019applying}, as the issues with conventional evaluation metrics remain.

%% file: chapter/04_methodology.tex
\begin{figure*}[th]
    \centering
    \includegraphics[width=0.7\textwidth]{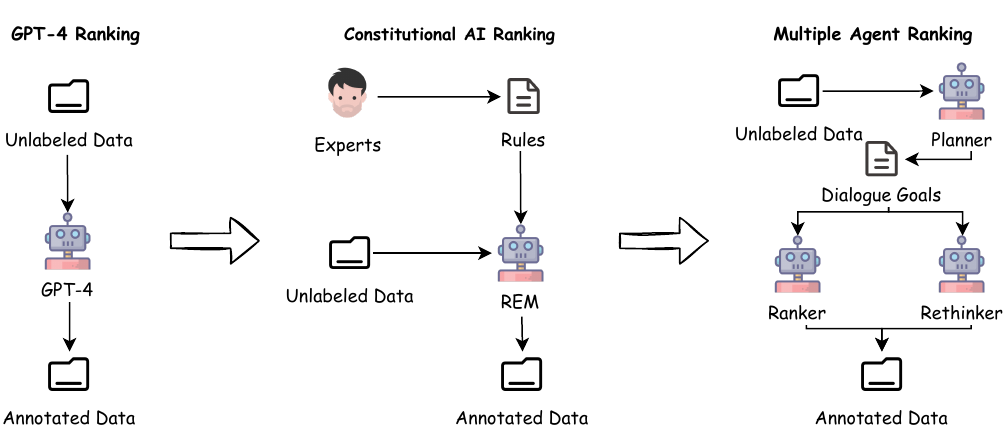}
    \caption{Three methods for labeling preference alignment data, where the strategy on the right improves upon the strategy on the left.}
    \label{fig:method}
\end{figure*}

\begin{figure}
    \centering
    \includegraphics[width=0.48\textwidth]{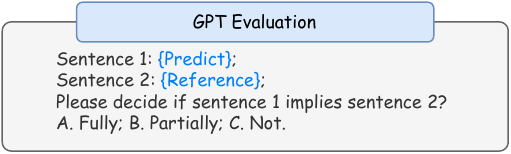}
    \caption{Prompt words for GPT-4 Evaluation.}
    \label{fig:gpt_eval_prompt}
\end{figure}

\section{Annotate Strategy Exploration}
\subsection{Overview}
We consider the annotation of preference alignment data as a candidate response ranking task. Traditionally, the RLAIF method uses a robust general-purpose LLM, such as GPT-4, for direct ranking. However, the effectiveness of this method in the medical field is debatable due to the significant differences between the preferences of GPT-4 and those of professional physicians. To address this gap between LLM and physician preferences, we analyze three data annotation strategies illustrated in Figure~\ref{fig:method}. The strategy on the right side of the figure represents an enhancement over the one on the left side. Specifically, in CAI Ranking, the model annotates preference data using rules established by human experts. Since these rules embed expert knowledge, the CAI ranking evaluation perspective aligns more closely with human experts than the GPT-4 ranking. In contrast, Multiple Agent Ranking (the method proposed in this paper) allows LLMs to create their own rules and set dialogue goals based on them. Unlike CAI Ranking, Multiple Agent Ranking does not need expert involvement, making it more adaptable. We elaborate on these three data annotation strategies in the upcoming subsections.

\begin{figure}[ht]
    \centering
    \includegraphics[width=0.48\textwidth]{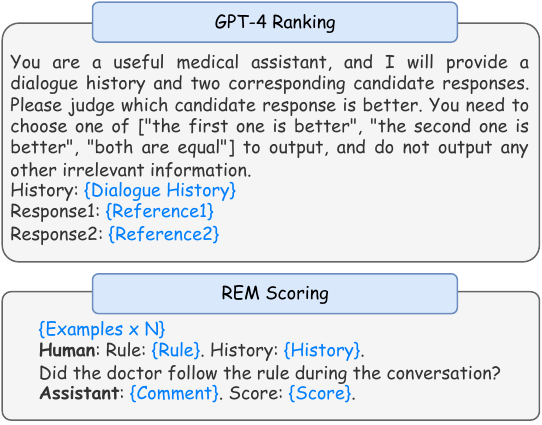}
    \caption{Prompt words utilized for GPT-4 Ranking and CAI Ranking are shown. GPT-4 Ranking employs a single-choice question to evaluate the candidate responses, whereas CAI Ranking uses a scoring technique. To avoid bias in GPT-4's comprehension of the rules, we apply In-Context Learning to improve the evaluation process, and the number of examples is five.}
    \label{fig:rp}
\end{figure}

\subsection{GPT-4 Ranking}
Currently, GPT-4 has been commonly used for the evaluation of medical dialogue models. This approach is primarily founded on the assumption that GPT-4's preferences closely resemble human preferences. To verify this assumption, we suggest employing GPT-4 to directly label a preference alignment dataset. The prompts used are displayed in Figure~\ref{fig:rp}~(GPT-4 Ranking). During the training of dialogue models using preference alignment data, we exclude data that GPT-4 considers as ``both are equal''.

\subsection{Constitutional AI Ranking}
The fundamental principle of Constitutional Artificial Intelligence (CAI) involves establishing a set of rules (referred to as a constitution) and assessing a model's output with these guidelines. Our study utilizes this framework for annotating data in the context of aligning medical preferences. As illustrated in Figure ~\ref{fig:method} (Constitutional AI Ranking), the annotation workflow comprises three stages: the development of rules, the construction of a rule evaluation model (REM), and data annotation. Here are the details of the process.

\begin{figure*}[th]
    \centering
    \includegraphics[width=0.8\textwidth]{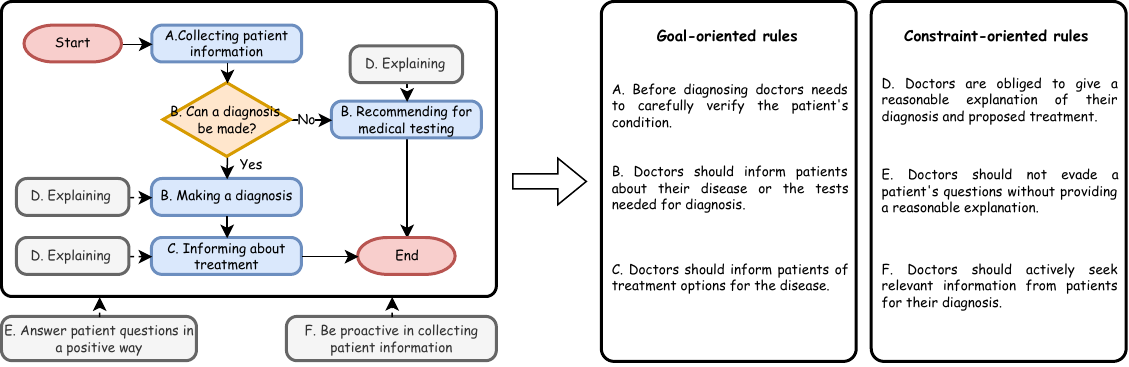}
    \caption{Medical diagnosis flowchart~(left) and the corresponding rules~(right). In the flowchart, we use blue boxes for activities, orange diamonds for judgment conditions, and gray boxes for additional constraints. We use the letters A-F to indicate the correspondence between the rules and the elements in the flowchart.}
    \label{fig:cai}
\end{figure*}

\subsubsection{Rules Development}
In the healthcare field, it is recommended that physicians adhere to clinically validated protocols for disease diagnosis to improve diagnostic accuracy. These protocols offer a systematic method for diagnostic procedures, supporting physicians in crucial decision making. Inspired by this approach, we argue that rule development should focus on enhancing the dialogue model's decision-making abilities. Using a format similar to a medical guideline, we use a flowchart, shown in Figure~\ref{fig:cai}, to represent the dialogue options of a doctor in an outpatient setting.

We categorize rules by their function into goal-oriented rules and constraint-oriented rules. Goal-oriented rules outline the goals that must be achieved for a full medical dialogue, whereas constraint-oriented rules define the limitations that the model must adhere to before achieving the goals. 

\subsubsection{Construct REM}
After establishing predefined rules, it is crucial to construct a REM for these rules. The evaluation task for the rules is organized in a Q\&A format, as illustrated in Figure~\ref{fig:rp} (REM scoring).
In this format, `Rule', `History', `Comment', and `Score' represent the slots to be filled. `Rule' and `History' serve as the model inputs, whereas `Comment' and `Score' are the outputs generated by the model. To streamline the scoring procedure, we assign the score values of 0, 1, and 2, which indicate noncompliance, partial compliance, and complete compliance with the rule, respectively.

We investigate two different approaches for establishing REM: firstly, we utilize In-context Learning techniques to direct GPT-4 in evaluating each candidate response. For this approach, we manually construct five examples per rule, covering all possible score ranges. The second approach involves annotating 1800 training instances to train the Baichuan-7B model. To reduce the labeling workload, we perform a rough labeling of these 1800 samples using the first method, follow by manual correction of the inaccuracies. Evaluation on 600 samples indicates comparable performance for both methods. We suspect that this is due to the relatively limited capability of the Baichuan2-7B model. Consequently, for the remainder of this work, we opt for the first construction approach.

\subsubsection{Data Annotation}
Given that the CAI ranking encompasses various rules, it is essential to integrate the scores of these diverse rules to rank the candidate responses. Our fundamental approach involves assigning distinct weights to each rule based on its significance and using a linear weighting technique to compute the total score of the candidate responses for ranking purposes. The scoring formula is as follows:
\begin{equation}
s(c \mid h) = \sum_{r}w_rs^r(c \mid h)
\label{eq:score}
\end{equation} 
In shich, $h$ represents the dialogue history, $c$ is the candidate response, $r$ is a specific rule, and $w_r$ is the weight assigned to that rule. The score $s$ is determined by REM.

We explore two distinct approaches for assigning weights to these rules, in order to better understand the method for labeling preference alignment data. The two methods are:
1) Average Weighting: This approach assigns equal weight to all rules, ignoring any dependencies among them.
2) Dependency-Based Weighting: This method accounts for the sequence of actions within the process. It assumes that if a preceding activity is not executed, the results of the subsequent activity are deemed unreliable. The weighting formula under this method is as follows:
\begin{equation}
    w_{r}=\prod_{r' \prec r} \mathcal{V}^{t_1}_{\alpha}(s^{r'}) \prod_{r'' \rightarrow r} \mathcal{V}^{t_2}_{\beta}(s^{r''}) 
\end{equation}
\begin{equation}
\mathcal{V}^t_y(x) = 
\begin{cases}
1 & x \ge t \\
y & \textit{otherwise}
\end{cases}
\end{equation}
Here, $r'$ represents a rule that precedes rule $r$, for example, in Figure~\ref{fig:cai}, rule $A$ precedes rule $B$. $r''$ indicates a rule that restricts $r$, such as Rule $D$ being a constraint for Rules $B$ and $C$. In our experiments, we assign $\alpha$ and $\beta$ to 0.1 and 0.8, respectively, suggesting a higher emphasis on achieving the goal rather than adhering to constraints. The thresholds $t_1$ and $t_2$ are set to 0.5.

In multiturn medical dialogues, it can be more effective to consider future rounds of doctor-patient interactions when ranking candidate responses. We represent the dialogue trajectory of future rounds as $(u_1, a_1, \ldots, u_i, a_i)$, where $u$ stands for the patient's response and $a$ stands for the model's response. Based on this, we propose a novel scoring function, with the following formula:
\begin{equation}
  s'(c \mid h) = s(c \mid h) + \sum_{i=1}^{n} d^i s(a_i \mid h, c, \ldots, u_i)
\end{equation}
Here, $n$ represents the maximum length of the trajectory of interest, and the discount factor $d \in (0, 1]$ reflects the importance assigned to future impacts. 

We label the higher-scored response as `Accepted' and the lower-scored response as `Rejected' for each data instance. It is observed that the greater the discrepancy between the scores of the two candidate responses, the more accurately the data are labeled. Hence, we exclude data with a score difference of less than one to maintain high data quality.

\subsection{Multiple Agent Ranking}
\begin{figure*}[th]
    \centering
    \includegraphics[width=0.9\textwidth]{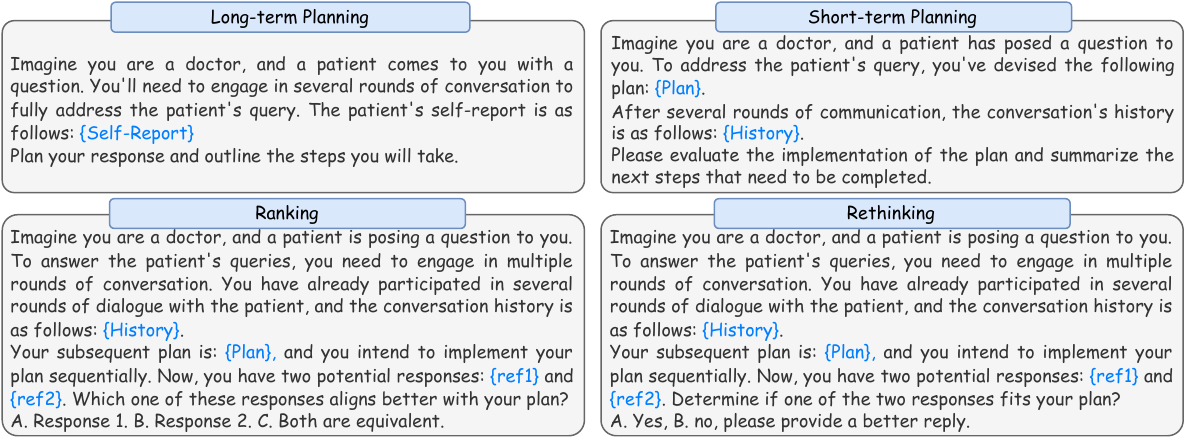}
    \caption{Prompts used by Multiple Agent Ranking. The Planner utilizes the prompts for Long-turn Planning and Short-turn Planning, whereas the Ranker and Rethinker make use of the prompts for Ranking and Rethinking respectively.}
    \label{fig:ap}
\end{figure*}

Our Multiple Agent Ranking method is inspired by CAI Ranking. We initially establish dialogue guidelines and then rank the candidate responses based on how well they adhere to these guidelines, as illustrated in Figure~\ref{fig:method} (Multiple Agent Ranking). Unlike CAI Ranking, Multiple Agent Ranking removes the need for expert involvement. Instead, it uses Planner to create guidelines from the dialogue history and set dialogue goals for candidate responses accordingly. The Ranker, similar to REM, directly ranks candidate responses based on these dialogue goals, avoiding complicated score aggregation. Additionally, if neither candidate response meets the criteria, Rethinker is used to generate a new response. Details of these three modules will be provided in the following sections.

\subsubsection{Planner}
We use GPT-4 to create the Planner, with the associated prompt phrases for this module shown in Figure~\ref{fig:ap} (Long-term planning, Short-term planning). In this context, long-term planning prompts help GPT-4 in generating a relevant dialogue guideline based on the patient's Self-Report, which consists of a series of specific dialogue steps. By doing this, we can avoid the need to develop a dialogue process similar to the one illustrated in Figure~\ref{fig:cai}, allowing GPT-4 to formulate a more personalized dialogue process based on the patient's symptoms.

After formulating a dialogue guideline, we transform it into immediate dialogue objectives suitable for subsequent stages. A short-term planning prompt is used to instruct GPT-4 in this task. In this prompt, the dialogue guideline and dialogue history are inserted into the 'Plan' and 'History' fields, respectively.

\subsubsection{Ranker and Rethinker}
The Ranker's task involves evaluating the data by determining which candidate response best meets the dialogue objective. Figure~\ref{fig:ap} (Ranking) presents the prompt we used. In this prompt, the dialogue objectives summarized by Short-term Planning are positioned in the 'Plan' slot.

During the annotation process with Ranker, we find that around 60\% of the data are tagged as 'Both are equivalent', resulting in a relatively small dataset. This problem stems from the conflict between the detailed plans of Planner and the doctors' practice of skipping some examination steps for diagnostic efficiency in real-world settings. Consequently, Ranker concludes that none of the candidate responses comply with the dialogue guidelines. However, we consider thorough check-up vital for the safety of medical dialogues. Therefore, we suggest the Rethinker module to create improved responses for those flagged as ``Both are equivalent''. The prompt words we used are shown in Figure~\ref{fig:ap} (Rethinking).

To clarify the annotation process of Rethinker, suppose that the data to be annotated comprise $(h, c_1, c_2)$, where $h$ represents the dialogue history, and $c_1$, $c_2$ are the candidate responses. Initially, Rethinker is used to generate an improved candidate response $c_3$, allowing us to create two preference-aligned samples: $(h, c_3 > c_1)$ and $(h, c_3 > c_2)$. However, our experiments revealed that the model tended to overfit $c_3$ during training, causing a drop in performance (refer to Section~\ref{sec:balance} for analysis). Consequently, we adopted the GPT-4 Ranking method to re-rank $c_1$ and $c_2$ to create a new sample. Therefore, for cases where the data is marked as ``Both are equal'', the Rethinker model generates three sets of annotated data. This approach has proven to significantly improve the guidance capabilities of the dialogue model.

%% file: chapter/05_exp_result.tex
\section{Experimental Results}
\begin{table*}[]
\caption{
Evaluation results for preference learning performance. The top scores are highlighted in bold. CSPT assesses the model's capability to guide users, whereas webMedQA assesses the model's instruction-following ability. Meddg and IMCS gauge the model's flexibility in utilizing user guidance and following instructions.
The SFT model is the base dialogue model described in Section~\ref{sec:backbone}. The CAI ranking with DEP is a Constitutional AI ranking method that considers rule interdependencies.
}
\label{tab:main_result}
\centering
\begin{tabular}{lccccccccc}
\toprule
\multicolumn{1}{l|}{\multirow{2}{*}{Model}} &
  \multicolumn{3}{c|}{CSPT} &
  \multicolumn{2}{c|}{webMedQA} &
  \multicolumn{2}{c|}{Meddg} &
  \multicolumn{2}{c}{IMCS} \\
\multicolumn{1}{l|}{} &
  \multicolumn{1}{c}{Sym. $\uparrow$} &
  \multicolumn{1}{c}{Test $\uparrow$} &
  \multicolumn{1}{c|}{Dis. $\uparrow$} &
  \multicolumn{1}{c}{R@L $\uparrow$} &
  \multicolumn{1}{c|}{GPD $\downarrow$} &
  \multicolumn{1}{c}{R@L $\uparrow$} &
  \multicolumn{1}{c|}{GPD $\downarrow$} &
  \multicolumn{1}{c}{R@L $\uparrow$} &
  \multicolumn{1}{c}{GPD $\downarrow$} \\ \midrule
\multicolumn{1}{l|}{SFT w/o alignment} &
  10.8 &
  26.3 &
  \multicolumn{1}{l|}{39.1} &
  19.6 &
  \multicolumn{1}{l|}{1.26} &
  27.0 &
  \multicolumn{1}{l|}{1.67} &
  19.5 &
  1.71 \\ \midrule
\multicolumn{10}{c}{\textit{Off-the-shelf Model}} \\ \midrule
\multicolumn{1}{l|}{Baichuan2-Chat~\cite{yang2023baichuan}} &
  21.7 &
  27.4 &
  \multicolumn{1}{l|}{33.8} &
  18.9 &
  \multicolumn{1}{l|}{1.44} &
  34.2 &
  \multicolumn{1}{l|}{1.69} &
  27.9 &
  1.77 \\
\multicolumn{1}{l|}{ChatGLM3~\cite{du2022glm}} &
  3.0 &
  23.6 &
  \multicolumn{1}{l|}{22.9} &
  22.7 &
  \multicolumn{1}{l|}{1.23} &
  26.1 &
  \multicolumn{1}{l|}{1.66} &
  20.5 &
  1.73 \\
\multicolumn{1}{l|}{DISC-MedLLM~\cite{bao2023disc}} &
  24.8 &
  34.9 &
  \multicolumn{1}{l|}{41.3} &
  24.7 &
  \multicolumn{1}{l|}{1.21} &
  28.9 &
  \multicolumn{1}{l|}{1.59} &
  25.0 &
  1.63 \\
\multicolumn{1}{l|}{Huatuo-II~\cite{huatuogpt-2023}} &
  5.7 &
  39.8 &
  \multicolumn{1}{l|}{40.7} &
  \textbf{28.1} &
  \multicolumn{1}{l|}{\textbf{1.11}} &
  33.2 &
  \multicolumn{1}{l|}{1.60} &
  37.3 &
  1.73 \\ \midrule
\multicolumn{10}{c}{\textit{RLAIF Model}} \\ \midrule
\multicolumn{1}{l|}{Gpt4 Ranking} &
  6.6 &
  42.2 &
  \multicolumn{1}{l|}{44.2} &
  22.9 &
  \multicolumn{1}{l|}{1.29} &
  37.7 &
  \multicolumn{1}{l|}{1.53} &
  33.9 &
  1.47 \\
\multicolumn{1}{l|}{CAI Ranking w DEP} &
  24.1 &
  41.1 &
  \multicolumn{1}{l|}{56.7} &
  21.5 &
  \multicolumn{1}{l|}{1.19} &
  34.8 &
  \multicolumn{1}{l|}{1.53} &
  25.4 &
  1.60 \\
\multicolumn{1}{l|}{Agent Ranking} &
  \textbf{32.1} &
  \textbf{58.6} &
  \multicolumn{1}{l|}{\textbf{67.5}} &
  24.2 &
  \multicolumn{1}{l|}{1.24} &
  \textbf{40.8} &
  \multicolumn{1}{l|}{\textbf{1.46}} &
  \textbf{37.5} &
  \textbf{1.39} \\ \bottomrule
\end{tabular}
\end{table*}

\subsection{Baselines and Training details}
We employ a variety of models as baselines, which can be categorized into two types: 1) Chat LLMs, containing ChatGLM3 (6B)~\cite{du2022glm,zeng2022glm} and Baichuan2-Chat (7B)~\cite{yang2023baichuan}; 2) Medical LLMs, including DISC-MedLLM~\cite{bao2023disc} and Huatuo-II~\cite{chen2023huatuogpt}. All medical models are trained using RLHF.

The chat model is trained with RLAIF on the foundational model outlined in Section \ref{sec:backbone}, utilizing four A100-40G GPUs. The training process incorporates the LoRA technique \cite{lora}, setting the parameter $\alpha$ and LoRA $r$ of LoRA to 16 and 64, respectively, and using a learning rate of 1e-4. Within the backbone model, we specifically focus on training the ``W\_pack'' and ``o\_proj'' modules, which denote the attention layer and projection layer of the Baichuan2-Chat model, respectively. The process is carried out with a batch size of 2, along with 16 gradient accumulation steps to efficiently handle memory and computation resources.

\subsection{Performance of Off-the-shelf Model}
The experimental findings are presented in Table ~\ref{tab:main_result}(Off-the-shelf Model). Our research demonstrates that specialized models tend to surpass generic dialogue models in healthcare-related dialogues. It is noted that no current model excels in both instruction following and user guidance effectively. For example, Huatuo-II shows a strong ability to follow instructions but is inadequate in the crucial user-guide task of effectively collecting patient symptom information. In contrast, DISC-MedLLM, although not as proficient as Huatuo II in following instruction, demonstrates superior performance in obtaining patient information, thus enhancing its diagnostic capabilities. Regarding information comprehensiveness, Huatuo-II outshines the other models, as reflected by its higher R@L score. When evaluating semantic understanding, DISC-MedLLM excels, exhibiting greater alignment in dialogue behaviors with those of medical professionals.

\subsection{Performance of GPT-4 Ranking}
Our analysis reveals notable differences in the effectiveness of various methods for labeling preference alignment data, as shown in Table~\ref{tab:main_result}~(RLAIF Model). GPT-4 Ranking improves the model's hybrid ability but does not greatly enhance individual skills. The primary improvement is observed in the model's ability to dispense medical advice, as evidenced by the substantial enhancement in the model's performance metrics on the R@L for the WebMedQA dataset. This improvement suggests that the model can more effectively encompass the content found in doctors' responses. Furthermore, the model demonstrates progress in the `Test' metrics of the CSPT, indicating a greater ability to provide more comprehensive recommendations for medical tests to patients.

However, this approach significantly reduces the model's ability to collect patient data during standardized patient assessments and achieves a similar poor performance to Huatuo-II on the `Dis.' metric of CSPT. The technical report of Huatuo-II~\cite{huatuogpt-2023} particularly notes that Huatuo-II's preferences are closely aligned with GPT-4 in answering medical questions. This suggests that a common problem is that following GPT-4's preferences may diminish the model's proficiency in guiding users.

\subsection{Performance of CAI and Agent Ranking}
The experimental results on CSPT indicate that both CAI Ranking and Agent Ranking markedly improve the user guidance capabilities of the model. However, it is noted that the CAI ranking provides the least effective hybrid performance among the three methods assessed. This is due to CAI Ranking's strict adherence to a predetermined workflow, as demonstrated in Figure~\ref{fig:cai}, which lacks the necessary flexibility for optimal functioning in various scenarios. In contrast, the GPT4 ranking and Agent Ranking do not follow a fixed protocol, allowing them to adapt more effectively to various situations. In particular, Agent Ranking's Planner tailors processes for each specific dialogue, highlighting its superior adaptability. 

Although Agent Ranking does not excel in instruction follow ability, according to the webMedQA evaluation results, it still demonstrates substantial improvement over the SFT model. We attribute this to the model's occasional difficulty in concurrently following instructions and guiding the user, necessitating a compromise. Thus, the Agent Ranking strategy, which aims to enhance overall performance, opts to slightly reduce its ability to follow commands to better guide users.

\begin{table*}[]
\caption{Preference learning performance evaluation results (Qwen Backbone). The highest scores are highlighted in bold.}
\label{tab:qwen_result}
\centering
\begin{tabular}{l|ccc|cc|cc|cc}
\toprule
\multirow{2}{*}{Model} & \multicolumn{3}{c|}{CSPT} & \multicolumn{2}{c|}{webMedQA} & \multicolumn{2}{c|}{Meddg} & \multicolumn{2}{c}{IMCS} \\
 & Sym.$\uparrow$ & Test$\uparrow$ & Dis.$\uparrow$ & R@L$\uparrow$ & GPD$\downarrow$ & R@L$\uparrow$ & GPD$\downarrow$ & R@L$\uparrow$ & GPD$\downarrow$ \\ \midrule
SFT w/o alignment & 15.9 & 30.1 & 33.8 & 18.5 & 1.26 & 26.4 & 1.60 & 18.5 & 1.71 \\
GPT4 Ranking & 14.5  & 41.1 & 47.9 & 23.6  & 1.35  & 39.8 & 1.51  & 37.1 & 1.51  \\
CAI Ranking w DEP & 25.9 & 44.1 & 54.4 & 23.8 & \textbf{1.14} & 36.9 & 1.56 & 28.0 & 1.59 \\
Agent Ranking & \textbf{40.3} & \textbf{57.2} & \textbf{63.7} & \textbf{24.0}  & 1.34  & \textbf{41.8}  & \textbf{1.35}  & \textbf{39.0} & \textbf{1.49}  \\ \bottomrule
\end{tabular}
\end{table*}

\subsection{Performance on Different Base Models}
To evaluate the effectiveness of our proposed RLAIF scheme across different base models, we perform experiments using the Qwen model. The results of these experiments are presented in Table~\ref{tab:qwen_result}. After testing, we observe that the Agent Ranking method consistently delivers excellent outcomes in all evaluations, thereby confirming the method's effectiveness. Additionally, we note that the intrinsic preferences of the base model itself influence RLAIF's efficacy to some extent. For example, before preference alignment, Baichuan-2 SFT has a relatively low symptom collection rate (10.8\%) but a higher correct diagnosis rate (39.1\%), while Qwen SFT exhibites a higher symptom collection rate (15.9\%) but a lower correct diagnosis rate (33.8\%). This indicates that the Qwen model is more proactive in symptom gathering but less effective in diagnosing symptoms compared to Baichuan-2 SFT. This difference persists even after applying RLAIF. Consequently, selecting models with preferences that align with the target preferences is crucial before executing RLAIF.

\begin{table}[htbp]
\caption{Diagnostic scores of Baichuan2-Chat and DISC-MedLLM without avoidance strategies.}
\label{tab:compare}
\centering
\scalebox{0.8}{
\begin{tabular}{@{}lcccc@{}}
\toprule
Model & Rejected Rate & Acc & Fixed Acc & Fixed CAI  \\ \midrule
Baichuan2-Chat & 31.9 & 33.7 & 49.7 & 54.8  \\
DISC-MedLLM & 20.9 & 41.3 & 54.1 & 56.6\\ 
\bottomrule
\end{tabular}
}
\end{table}

\subsection{Impact of Diagnostic Avoidance on Correct Diagnostic Rates}
Merely collecting more information does not inherently improve diagnostic capabilities. Baichuan2-Chat and DISC-MedLLM, while strong in information gathering, exhibit limited diagnostic effectiveness. The main cause of this limitation is that these models often avoid making diagnoses for safety reasons. To further investigate the effect of this avoidance strategy on diagnostic accuracy, we determine the scores that Baichuan2-Chat and DISC-MedLLM have attained without using the avoidance strategy, known as Fixed Acc. Concurrently, to guarantee an equitable comparison, we also calculate the CAI Ranking scores using the same data subsets as those used for the two models, called Fixed CAI here. 

The results are shown in Table~\ref{tab:compare}. DISC-LLM demonstrated diagnostic accuracy comparable to CAI Ranking, while Baichuan2-Chat displayed somewhat lower accuracy due to inadequate fine-tuning on the medical dataset. Moreover, Fixed CAI scores are quite close to the CAI Ranking scores shown in Table~\ref{tab:main_result}. Considering that all models are Baichuan2-based, diagnostic capabilities are not significantly different, implying that the avoidance strategies adopted by Baichuan2-Chat and DISC-MedLLM do not effectively minimize the risk of misdiagnosis.

%% file: chapter/06_discussion.tex
\section{Discussion}
\label{sec:discuss}
In the previous section, our experiments demonstrate that Agent Ranking surpasses all other strategies in overall performance, markedly outperforming existing open source models. To further clarify these outcomes, we examine three principal factors: the distribution of preferences across the three strategies, the influence of long-term planning, and the balance between user guidance and instruction following abilities.

\subsection{The Preference Distribution of Three Strategies.}

\begin{figure*}[]
\centering
\includegraphics[width=0.9\textwidth]{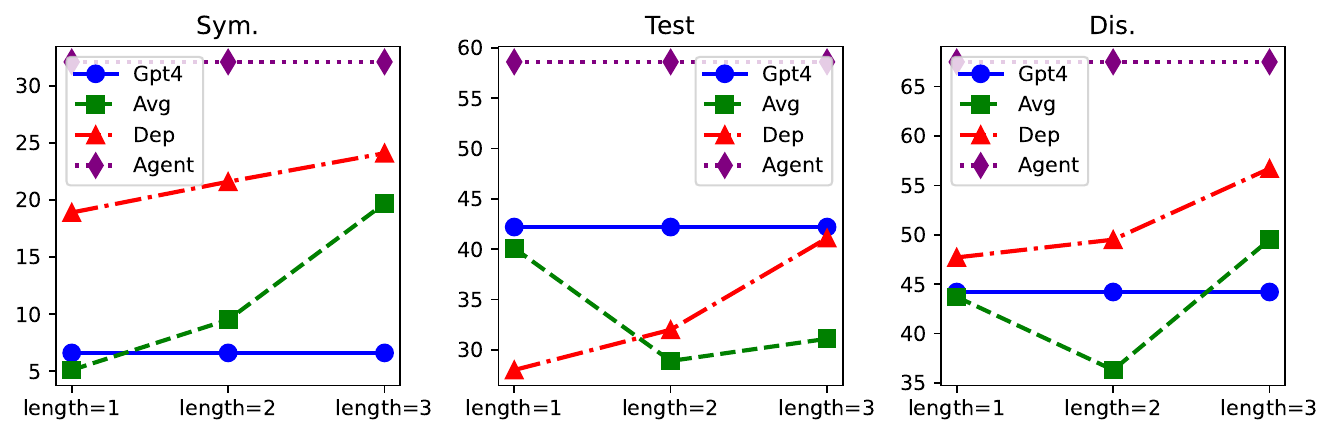}
\caption{The performance comparison between the Avg strategy and the Dep strategy is influenced by the Trace Length, which represents the length of future dialogues (trajectories) considered. A length of 1 indicates that only the current round of dialogue is considered, without taking into account the impact of future interactions. Due to the influence of the inherent length of dialogues in the dataset, we have set the upper limit of Trace Length to 3.}
\label{fig:ab_study}
\end{figure*}

\begin{figure}
    \centering
    \includegraphics[width=0.4\textwidth]{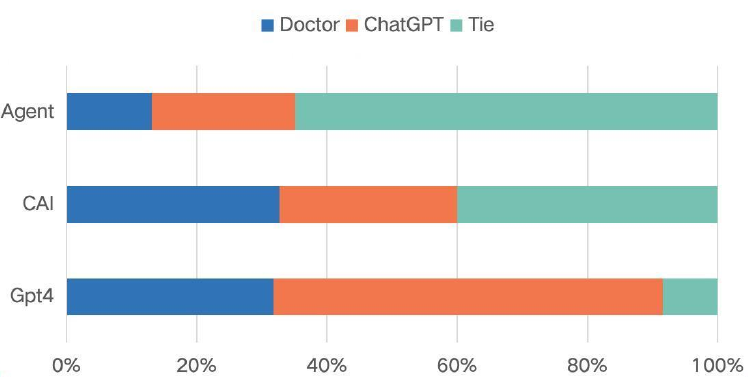}
    \caption{Preferences of different annotation strategies for candidate responses, where Tie indicates that there is no obvious preference tendency for the two candidate samples.}
    \label{fig:compare}
\end{figure}

To examine the differences in the efficacy of the three annotation strategies, a simple approach is to analyze their preference distributions. For this analysis, we conducted statistical evaluations, with the results presented in Figure~\ref{fig:compare}. Our findings indicate that GPT4 Ranking favors responses produced by ChatGPT, which typically adopts a passive communication style and does not actively question users. This suggests that models trained with GPT4 Ranking do not actively seek information from patients. Conversely, CAI Ranking favors responses from doctors, which tend to be more proactive; as a result, models trained with CAI Ranking are better at guiding users.

While Agent Ranking also leans towards ChatGPT's output, it is noteworthy that the Rethinker module generates superior responses for a significant portion of data marked as Tie. These superior responses often involve unsolicited requests for patient information, demonstrating that Agent Ranking leads to the development of a model with the strongest capacity for effective user guidance.

An interesting finding is that Agent Ranking and CAI Ranking annotate many samples as Tie. Furthermore, the effectiveness of the trained dialogue model improves as the ratio of this data segment increases. We determine that this occurs due to the integration of a pre-established dialogue flow within the annotation procedure. A substantial portion of the data contains two candidate responses that do not meet the predefined criteria and are consequently labeled as Tie. This observation implies that maintaining correct processes is crucial for aligning medical preferences.

\subsection{The Impact of Long-turn Planing.}
Our study shows that Agent Ranking and CAI Ranking substantially exceed GPT-4 Ranking in diagnosis within the CSPT structure. Both Agent Ranking and CAI Ranking begin by outlining a long-term conversational guideline and assess responses based on their adherence to this pathway. However, GPT Rank assesses whether a response addresses the patient's immediate issue, disregarding its future implications.

To improve our understanding of how long-term planning influences preference alignment, we conducted an ablation study, the results of which are shown in Figure~\ref{fig:ab_study}. This investigation focuses on two pivotal elements that impact the model's proficiency in executing long-term strategies: (1) the consideration of the sequence in which goals are achieved, and (2) the inclusion of future consequences. In Figure~\ref{fig:ab_study}, the term ``Avg'' denotes an approach where all goals are weighted equally, contrasting with ``Dep,'' which signifies the acknowledgment of a sequential dependency among goals during the CAI (Conversational Artificial Intelligence) Ranking process. The variable ``Length'' refers to the extent of future dialogue trajectories contemplated by the model~(see Section~\ref{sec:dp}); a Length of 1 implies the model's disregard for the repercussions of present dialogues on future interactions. In particular, the Agent Ranking and GPT-4 Ranking methodologies do not incorporate future dialogue trajectories from the dataset, rendering their scores invariant to changes in length. From this study, we discern two critical insights.

\paragraph{Finding-I}
When disregarding the sequence of goal completion and the influence on future dialogues, CAI Ranking and GPT-4 Ranking exhibit similar performance. 
Our observations show that Avg and GPT-4 Ranking are equally effective for CSPT when length = 1. Altering CAI Ranking's rule combination strategy from Avg to Dep leads to notable variations in the model's symptom collection and test recommendation abilities, implying that leveraging rule (goal) dependency can effectively modulate GPT-4's inclination preferences. Additionally, we detected changes in Avg's scores on CSPT dataset with incremental increases in Length, highlighting the necessity to account for the current dialogue's impact on forthcoming interactions.

\paragraph{Finding-II} 
Taking future effects into account will improve the model's ability to gather symptoms, but not its diagnostic recommendation accuracy. 
We observe that as Length increases, both AVG and DEP strategies show improvement in the ``Sym.'' metric, indicating that incorporating future impacts enhances the model's guidance ability. In contrast, a longer length causes significant variability in the ``Test'' metric for both AVG and DEP, neither exceeding the GPT-4 ranking score. This is likely because recommending medical tests is easier than collecting symptoms, which requires only a few symptoms for a reasonable test suggestion. The increased length also alters the model response patterns, affecting its performance in the ``Test'' metric. It is noted that Agent Ranking significantly exceeds other methods in test recommendations, attributed to the robust responses from the Rethinker module, which make the dialogue model's answers more thorough and substantially boost the ``Test'' score.

\begin{table}[h]
    \caption{Comparison of GPT4 Ranking, Rethinker, and Rethinker + GPT4 Ranking on various metrics}
    \label{tab:comparison}
    \centering
    \scalebox{0.9}{
    \begin{tabular}{lccc}
        \toprule
        Metrics & GPT4 & Rethinker & Rethinker+GPT4 \\ \midrule
        DPO Loss & 0.61 & 0.19 & 0.33 \\
        CSPT Sym. $\uparrow$ & 6.6 & 54.5 & 32.1 \\
        webMedQA GPD $\downarrow$ & 1.29 & 1.46 & 1.24 \\
        Hybrid GPD $\downarrow$ & 1.5 & 1.49 & 1.42 \\ \bottomrule
    \end{tabular}
    }
\end{table}

\subsection{The Balance between User Guidance and Instruction Following Abilities}
\label{sec:balance}

Individuals often encounter difficulties in simultaneously providing guidance and adhering to instructions, necessitating a balance between these competencies within a model. Our Agent Ranking approach integrates the Rethinker module to enhance the model's capability to offer guidance, as evidenced in Table~\ref{tab:comparison}(CSPT Sym). However, our analysis reveals that training the model solely with the data generated by the Rethinker compromises its instruction-following capacity, as illustrated in Table~\ref{tab:comparison} (webMedQA GPD). This decrease is due to the dominant presence of sentences aimed at guiding users in Rethinker-generated responses, which stylistically differ from the candidate responses in the original dataset. Consequently, the model becomes overly accustomed to the Rethinker-generated responses during training, significantly reducing its ability to follow instructions, as demonstrated in Figure~\ref{tab:comparison} (DPO loss).

To counteract this issue, a logical strategy involves enhancing the likelihood of generating candidate responses by the model. Specifically, we propose re-ranking the candidate responses labeled as ``Both are equal'' using GPT-4 Ranking. This method efficiently prevents overfitting by not excessively lowering the probability of candidate responses during the preference alignment process. Our findings suggest that while this intervention may slightly decrease the model's guidance capability, it substantially increases its overall performance, as depicted in Figure~\ref{tab:comparison}(Hybrid GPD).

%% file: chapter/07_conclusion.tex
\section{Conclusion}
In this paper, we focus on the problem on how to generate training data to better align doctors' preference in LLM-based medical dialogue generation and make three contributions. Firstly, we design a subjective evaluation system for LLM-based medical dialogue generation.
Secondly, we discover that the flow graph is good at descripting the doctor's preference.
Thirdly, we propose a multi-agent labeling strategy based on two aspects, i.e., user guidance and instruction following. 
Our experimental results validate the effectiveness of our multi-agent method in modeling the procedure correctness and its superiority on generalization to diverse scenarios.
In future, we will further investigate how to generate preference data to eliminate hallucinations of LLMs and process reject-answer situations. 

\section{Ethics Statement}
The datasets used in this study are open source and widely used. Our study relies exclusively on these datasets and is conducted in a simulated environment. The purpose of our study is to evaluate and analyze the impact of different RLAIF strategies on these datasets and does not involve the production of commercially available models of medical dialogue. Among these datasets, two warrant particular attention, for which we offer the following clarifications.

\begin{itemize}
    \item Medical Dialogue-CN~\cite{zeng2020meddialog}: This dataset, originating from Haodf.com and copyrighted by the platform, was partially utilized in this study. The subset used was extracted by the Disc-MedLLM~\cite{bao2023disc} project at Fudan University and rewritten and desensitized using ChatGPT.

    \item CSPT: This dataset was developed from the medical textbook "Objective Structured Clinical Examination \& Standardized Patients" published by People's Medical Publishing House, China~\cite{sp_testing}, with all samples undergoing desensitization.
\end{itemize}
- 

- 